\documentclass[letterpaper, 10 pt, conference, final]{ieeeconf}  

\IEEEoverridecommandlockouts                              

\usepackage{amsmath} 
\usepackage{amssymb}
\usepackage{xcolor}
\usepackage{graphicx}
\usepackage{float} 
\usepackage{marginnote}
\usepackage{multirow}
\usepackage{makecell}
\usepackage{subcaption}
\usepackage{soul}
\usepackage{booktabs}
\usepackage[T1]{fontenc}

\newcommand{\squishlist}{
	\begin{list}{$\bullet$}
		{ \setlength{\itemsep}{0pt}
			\setlength{\parsep}{1pt}
			\setlength{\topsep}{1pt}
			\setlength{\partopsep}{0pt}
			\setlength{\leftmargin}{1em}
			\setlength{\labelwidth}{1em}
			\setlength{\labelsep}{0.5em} } }
\newcommand{\squishend}{\end{list} 
}

\title{\LARGE \bf
LatentMimic: Terrain-Adaptive Locomotion via Latent Space Imitation
}
\author{
Zhiquan Wang, Yunyu Liu, Dipam Patel, Ayush Kumar, Aniket Bera, Bedrich Benes\\
{\textit{Department of Computer Science, Purdue University, USA}}\\
 \thanks{}{\texttt{\{wang4490, liu3154, dipam, kumar987, aniketbera, bbenes\}@purdue.edu}}
}
\IEEEoverridecommandlockouts  

\begin{document}

\maketitle
\pagestyle{plain}
\pagenumbering{arabic}

\begin{abstract}
Developing natural and diverse locomotion controllers for quadruped robots that can adapt to complex terrains while preserving motion style remains a significant challenge. Existing imitation-based methods face a fundamental optimization trade-off: strict adherence to motion capture (mocap) references penalizes the geometric deviations required for terrain adaptability, whereas terrain-centric policies often compromise stylistic fidelity. We introduce LatentMimic, a novel locomotion learning framework that decouples stylistic fidelity from geometric constraints. By minimizing the marginal latent divergence between the policy's state-action distribution and a learned mocap prior, our approach provides a conditional relaxation of rigid pose-tracking objectives. This formulation preserves gait topology while permitting independent end-effector adaptations for irregular terrains. We further introduce a terrain adaptation module with a dynamic replay buffer to resolve the policy's distribution shifts across different terrains. We validate our method across four locomotion styles and four terrains, demonstrating that LatentMimic enables effective terrain-adaptive locomotion, achieving higher terrain traversal success rates than state-of-the-art motion-tracking methods while maintaining high stylistic fidelity.
\end{abstract}

\section{Introduction}
Quadrupedal locomotion remains a fundamental challenge in robotics, particularly in enabling robots to execute agile, natural movements across diverse terrains while maintaining stylistic versatility. Designing controllers that incorporate multiple locomotion skills traditionally demands a labor-intensive process of reward engineering and parameter tuning. A prevalent approach to obtaining natural motion priors involves motion capture (mocap). However, mocap data are typically constrained to a single modality, such as walking on a flat surface, and inherently lack the geometric variations required for generalization across diverse terrains.

Recent advances in imitation learning have improved robotic locomotion by enabling robots to mimic natural motion styles (e.g., animals). However, they are limited in their ability to generalize beyond the constraints of their mocap data. Existing adversarial imitation methods enforce exact kinematic matching, which introduces a fundamental optimization conflict: strict adherence to reference kinematics penalizes the geometric deviations (e.g., increased foot clearance) necessary for terrain adaptability. Thus, effectively transferring learned locomotion skills to unseen terrains is pivotal for achieving versatile, real-world robotic mobility.  

Prior work in locomotion control has made significant strides by approximating the system dynamics and solving for optimal actions via optimization~\cite{Winkler2018-iz, Byl2009-am, Chi2022-nn, Semini2011-po, Xie2020-wx, Dario-Bellicoso2017-oc}. However, generalizing a singular model-based control strategy across multiple motion styles and diverse terrains remains computationally prohibitive. More recently, deep reinforcement learning (DRL) has demonstrated the potential to synthesize agile locomotion while enhancing generalization across terrains~\cite{Rudin2021-qd, Margolis2022-kc, Hwangbo2019-jc, Heess2017-mq,stkepien2025latent,lee2020learning,zhu2026transterrainawarereinforcementlearning,huang2026trainingsimulationquadrupedalrobot}. However, pure RL-based approaches often require extensive manual effort in reward engineering and hyperparameter tuning. Moreover, the learned policies frequently converge to kinematically suboptimal gaits that deviate from biological locomotion.

To synthesize biological gaits, recent works integrate motion references and imitation learning into the reinforcement learning objective~\cite{Peng2020-qm, Li2023-ru}. While these tracking-based methods effectively replicate reference behaviors on flat ground, they suffer from a rigid coupling between the reference locomotion style and the specific terrain geometry. As these works enforce exact kinematic matching, achieving skill generalization requires exhaustive, terrain-specific motion-capture data~\cite{Zhang2018-dd}.
However, collecting diverse motion references from real-world complex terrains is highly impractical, leaving the challenge of terrain-agnostic style imitation unsolved.

\begin{figure}[hbt]
\centering    \includegraphics[width=0.99\linewidth]{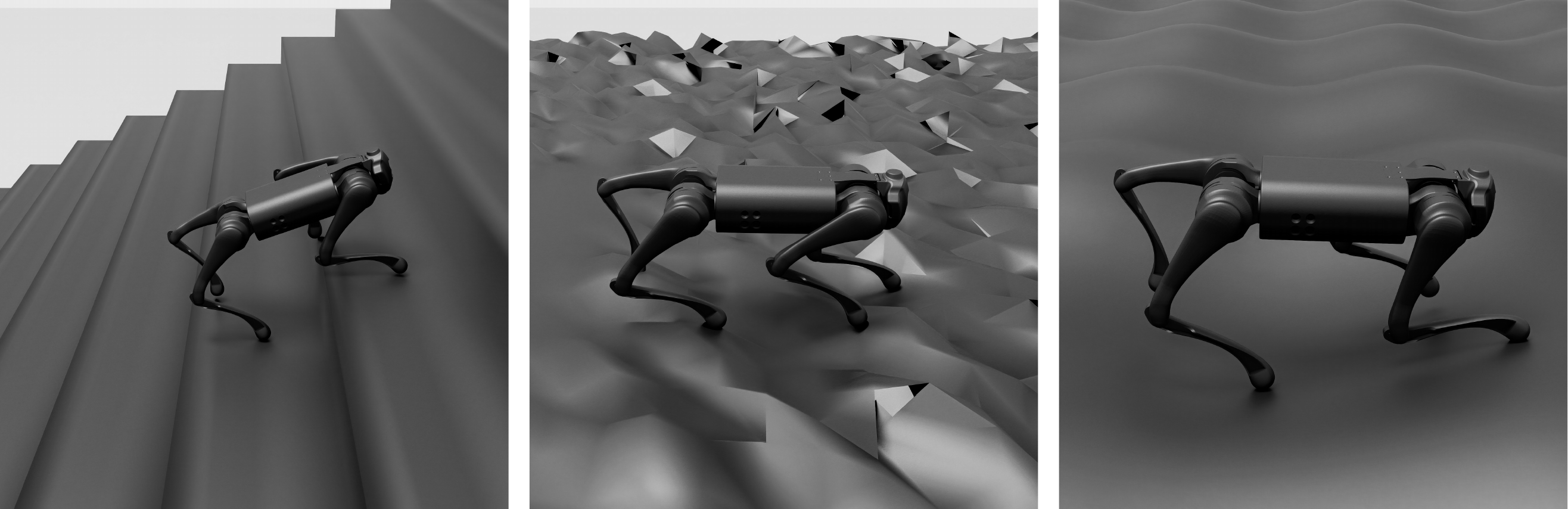}\caption{Left to right: Stair terrain, Noisy terrain, Wavy terrain.}\label{fig:terrains}\vspace{-3mm}
\end{figure} 
\begin{figure*}[hbt]
\centering
\includegraphics[width=.99\linewidth]{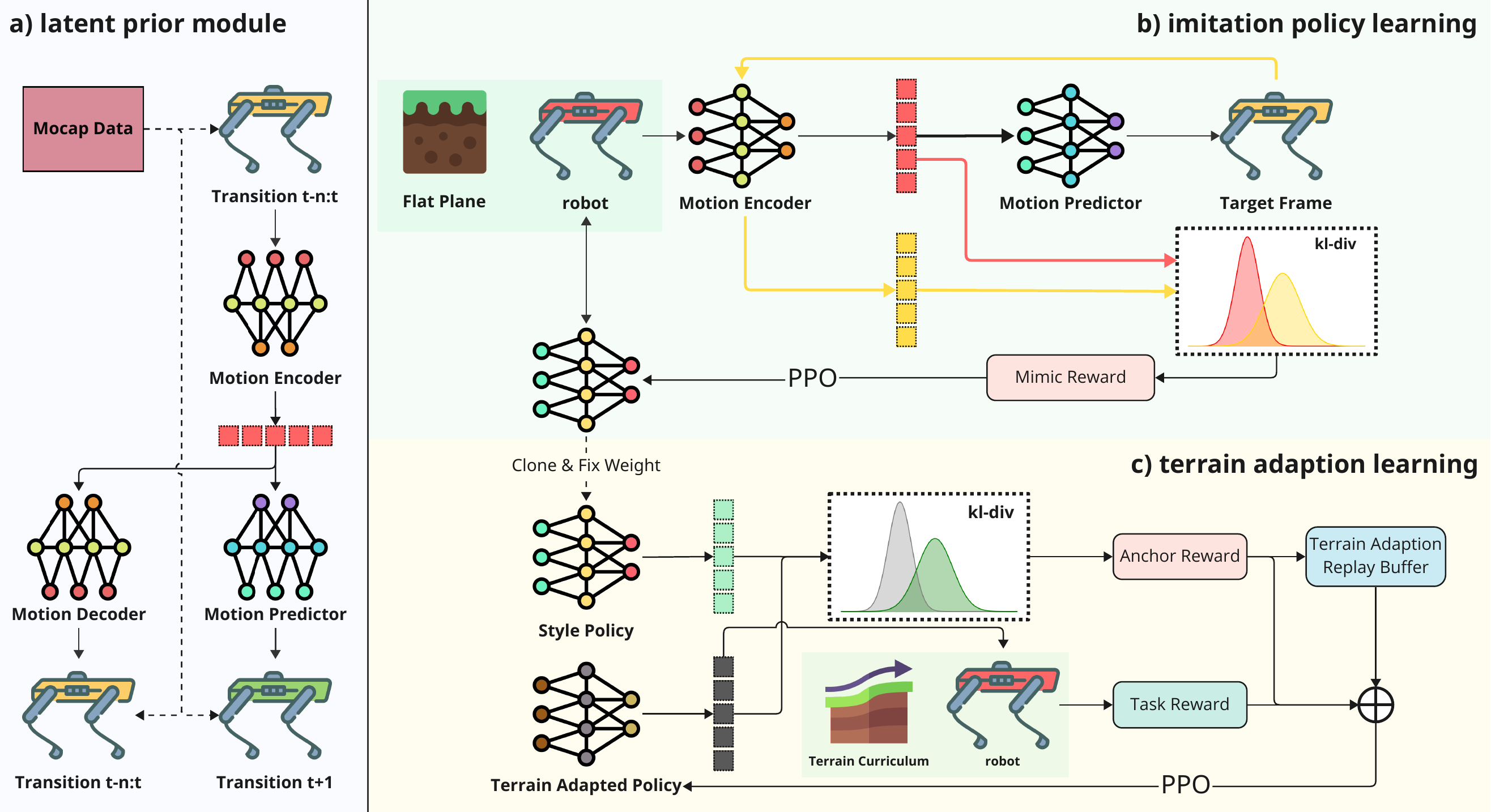}
    \caption{Overview of LatentMimic: a) We first pretrain a motion encoder to encode the motion transitions into a latent space and a motion predictor to predict the future motion transitions based on a mocap dataset. These two networks together form our \textbf{Latent Prior Module}. b) We then perform imitation policy learning using a mimic reward that measures the KL-divergence between the simulated and target transition in the latent space to learn the desired motion style. c)~We introduce an anchor reward and a terrain adaptation replay buffer to enhance terrain traverse while preserving the desired motion style. These components aid learning and enable effective adaptation across diverse terrains. Note: The latent prior module is trained in both b) and c).}
    \label{fig:overview}
    \vspace{-7mm}
\end{figure*}

We propose \textbf{LatentMimic}, a novel learning framework that enables imitation learning through a latent-space representation while explicitly decoupling stylistic fidelity from geometric constraints. As illustrated in Fig.~\ref{fig:overview}, we begin with motion capture data of a dog walking on flat terrain (a) and retarget the motion to the robot's morphology (b). Instead of enforcing exact kinematic matching in the observation space, we train the control policy by minimizing the marginal latent divergence between the reference and simulated motions. Crucially, this latent-space reward provides a conditional relaxation: it ensures the robot preserves the topological locomotion style while avoiding penalties for necessary geometric adaptations (e.g., altering joint angles to clear stairs) in the physical space. Subsequently, the robot is trained to traverse various terrains. Because executing these geometric adaptations on non-flat terrain inherently induces state distribution shifts, we introduce a terrain adaptation module that uses a dynamic replay buffer (c). By storing successful transitional motion frames, this buffer continuously updates the support set of the latent prior, ensuring that the target kinematics remain valid across complex geometries. Our main contributions are as follows:

\begin{enumerate}
    \item \textbf{A two-stage training pipeline} that enables robots to learn natural locomotion styles while generalizing across various terrains without requiring terrain-specific motion data.
    \item \textbf{A latent mimic reward} that promotes imitation of high-level locomotion styles rather than directly copying joint configurations or relying on extensive handcrafted rewards.
    \item \textbf{A latent motion adaptation module} that leverages incremental motion frames to enhance locomotion capability across different terrains and robustly resolve state distribution shifts.
\end{enumerate}

By integrating these theoretical mechanisms, our approach synthesizes robust quadrupedal locomotion that generalizes beyond the kinematic constraints of flat-ground reference data, marking a highly scalable methodology for the real-world deployment of agile robotic quadrupeds.

\section{Related Work}
\textbf{Deep Reinforcement Learning for Locomotion:} Model-free DRL facilitates the acquisition of versatile quadrupedal locomotion policies mapping proprioceptive states to joint commands~\cite{Rudin2021-qd, lee2020learning, Miki2022learning, Kumar_2021, Siekmann2021sim}. However, these methods exhibit strong dependencies on empirical reward engineering. Suboptimal scalar reward formulations frequently yield kinematically inefficient behaviors that deviate from biological norms~\cite{Miki2022learning, Siekmann2021sim, Lee2019robust, Matiisen_2020}. Furthermore, modulating distinct locomotion styles within a singular policy via scalar rewards remains mathematically under-constrained, which couples stabilization objectives with stylistic fidelity.

\textbf{Motion Imitation and Style Learning:} To bypass exhaustive reward tuning, reference-guided methods incorporate motion capture (mocap) data into the RL objective. Approaches minimizing state-tracking errors successfully synthesize biological movements~\cite{koenemann2014real, Peng_2018, Peng2020-qm,huang2025learning}. Furthermore, Adversarial Motion Prior (AMP)~\cite{Peng2021-ch} applies Generative Adversarial Imitation Learning (GAIL)~\cite{ho2016generative} to enforce kinematic distributions resembling the reference data, producing animal-like gaits~\cite{alej2022adversarial,Escontrela2022-hz, Wu2023-rt, Yang2023-jh, Wu2023-bb,stkepien2025latent,peng2025bcamp}. Nonetheless, because these techniques enforce exact kinematic matching, their generalization is strictly bounded by the geometric constraints of the source data (typically planar surfaces).

\textbf{Terrain Adaptation:} Navigating unstructured terrains introduces geometric variations that exceed the support distribution of flat-ground priors. Curriculum learning progressively scales terrain complexity~\cite{Rudin2021-qd, Margolis2022-kc, lee2020learning,zhu2026transterrainawarereinforcementlearning,huang2026trainingsimulationquadrupedalrobot}, while Rapid Motor Adaptation (RMA)~\cite{Kumar_2021} and Terrain-Aware Locomotion (TAL)~\cite{shi2023terrain} utilize history encoders or exteroceptive signals to estimate environmental extrinsics. To combine stylistic fidelity with adaptability, recent tracking approaches integrate terrain-specific mocap references~\cite{Li2023-ru}. However, relying on explicitly paired terrain-motion datasets restricts scalability. In contrast, LatentMimic explicitly decouples topological style from geometric constraints, enabling terrain-agnostic imitation using only flat-ground priors and a dynamic state-shift adaptation module, thereby eliminating the dependence on exteroceptive sensors or terrain-specific reference trajectories.

\section{Methodology}\label{s:methodology}
The LatentMimic framework is structured into four stages (Fig.~\ref{fig:overview}): (1) retargeting reference motion capture data via inverse kinematics; (2) pre-training a latent prior module to embed physical states into a latent space $\mathcal{Z}$ and forecast future frames; (3) optimizing a control policy by minimizing the latent divergence between simulated ($z_{\text{sim}}$) and target ($z_{\text{target}}$) representations; and (4) executing dynamic terrain adaptation. We first formalize how this latent-space formulation mathematically resolves the optimization conflicts inherent in full-kinematic adversarial methods~\cite{Peng2021-ch, alej2022adversarial}.
\subsection{Motivation and Problem Formulation}\label{sub:motivation}

Let $\mathcal{X} \subset \mathbb{R}^n$ denote the full kinematic space of the motion reference (e.g., base poses, joint angles, velocities, and end effector positions), and let $\mathcal{Z} \subset \mathbb{R}^d$ denote the learned latent representation space. We define $P_{ref}$ as the probability measure of the state transitions from the reference motion capture data, and $P_\pi$ as the probability measure generated by the control policy $\pi$. Specifically, these measures quantify the probability density of the robot exhibiting a particular kinematic and dynamic state configuration within $\mathcal{X}$ during locomotion.

Existing Adversarial Motion Prior (AMP) \cite{Peng2021-ch} methods explicitly train a discriminator as a reward signal to minimize the divergence between the simulated and reference distributions in the full kinematic space. It utilizes an $f$-divergence, $\mathbb{D}_f(P_{ref}(X) \parallel P_\pi(X))$, such as the Jensen-Shannon or Pearson $\chi^2$ divergence. Following established analytical methods in generative modeling \cite{chen2016infoganinterpretablerepresentationlearning, hu2018deepgenerativemodelslearnable, xiao2022stabilityanalysisgeneralizationbounds, nowozin2016fgantraininggenerativeneural}, we examine the Kullback-Leibler (KL) divergence as a representative objective to expose the underlying optimization conflict during distribution matching across hierarchical spaces, then the adversarial objective can be formulated as:
\begin{equation}
    \mathcal{J}_{AMP}(\pi) \approx \mathbb{D}_{KL}\left(P_{ref}(X) \parallel P_\pi(X)\right)
\end{equation}
By introducing a fixed encoder mapping $q(z|x)$ that projects the observation space into the latent space, and applying the chain rule for KL divergence, we can analytically decompose the full-kinematic objective into two distinct terms:
\begin{equation}
\begin{aligned}
    \mathcal{J}_{AMP}(\pi) &= \underbrace{\mathbb{D}_{KL}\left(P_{ref}(Z) \parallel P_\pi(Z)\right)}_{\substack{\text{Marginal Latent} \\ \text{Divergence}}} \\
    &\quad + \underbrace{\mathbb{E}_{Z}\left[\mathbb{D}_{KL}\left(P_{ref}(X|Z) \parallel P_\pi(X|Z)\right)\right]}_{\substack{\text{Conditional Geometric} \\ \text{Divergence}}}
\end{aligned}
\end{equation}

This fundamental decomposition exposes the optimization conflict in terrain-adaptive imitation. The second term, the expected conditional divergence with respect to $Z \sim P_{ref}(Z)$, enforces strict geometric alignment. It strictly penalizes any deviation in the physical observation space $X$ given a specific motion style $Z$. When navigating various terrains, such as stairs or uneven ground, physical multi-body dynamics require the robot to adjust its foot trajectories (e.g., increasing foot clearance). Consequently, the policy's conditional distribution $P_\pi(X|Z)$ must fundamentally deviate from the flat-ground reference $P_{ref}(X|Z)$. In this scenario, the conditional geometric divergence sharply increases, causing the AMP objective to penalize the exact geometric adaptations necessary for terrain survival.

To resolve this conflict, LatentMimic explicitly truncates the conditional geometric divergence from the objective. We formulate our imitation objective solely based on the marginal latent divergence:
\begin{equation}
    \mathcal{J}_{LM}(\pi) = \mathbb{D}_{KL}\left(P_{ref}(Z) \parallel P_\pi(Z)\right).
\end{equation}\label{eq:kl_div}\vspace{-4mm}

By the data processing inequality, $\mathcal{J}_{LM}(\pi) \le \mathcal{J}_{AMP}(\pi)$, establishing our objective as a strict lower bound. It ensures that the topological essence of the gait (e.g., phase variations and footfall sequences) is strictly preserved within the latent manifold $\mathcal{Z}$, while permitting necessary and unpenalized geometric deviations in the physical observation space $\mathcal{X}$. The subsequent modules of our framework are designed to implement this relaxed objective and dynamically resolve the out-of-distribution state transitions induced by these geometric adaptations.

\subsection{Motion Capture Data and Preprocessing}\label{sub:mocap}
We utilize a public mocap dataset~\cite{Zhang2018-dd} that captures the locomotion of a real dog to construct the motion reference distribution $P_{ref}(X)$. Given that the dataset was captured exclusively on flat terrain and lacks terrain-specific details (e.g., leg clearance required for stairs), and considering the morphological differences between the dog and our robot, the data are retargeted using inverse kinematics~\cite{Gleicher1998-iz}. Following the methodology in~\cite{Peng2020-qm}, the mocap dataset $M$ is represented as a sequence of frames, where each frame at time step $t$ is defined as
\begin{equation}
   m_t = \left(p_t, \theta_t, v_t, q_t, \dot{q}_t\right),
\end{equation}
with $p_t \in \mathbb{R}^3$ denoting the base position, $\theta_t \in \mathbb{R}^4$ representing the base orientation (quaternion), $v_t \in \mathbb{R}^6$ capturing the base velocities, and $q_t,\,\dot{q}_t \in \mathbb{R}^n$ representing the joint angles and joint velocities, respectively.

\subsection{Latent Prior Module}\label{sub:latentprior}
To optimize the marginal latent divergence objective defined in Eq. \ref{eq:kl_div}, we first establish the latent space $\mathcal{Z}$ and the projection mapping $q(z|x)$. Since directly matching flat-ground key frames on complex terrains can cause geometric conflicts, this module maps physical states to a latent space. Unlike previous methods that enforce strict tracking of key frames~\cite{Peng2020-qm,Zhang2018-dd}, this projection allows the underlying locomotion styles to be evaluated independently of local geometric variations.

\subsubsection{Motion Encoder}\label{ssub:mencoder}
The motion encoder ($E$) implements the mapping $q(z|x)$ that maps motion attributes into the latent space. Initially trained on the mocap data, it defines the locomotion style at time step $t$ over $w$ preceding frames:
\begin{equation}
    S_t^{\text{motion}} = \left(m_{t-w-1}, \ldots, m_{t-1}, m_t\right).
\end{equation}

The motion encoder is trained to reconstruct $S_t^{\text{motion}}$ in an autoencoder framework. The process of encoding the motion is formulated as:
\begin{equation}
    S_t^{\text{motion}} \rightarrow z_t \rightarrow \hat{S}_t^{\text{motion}}.
\end{equation}

We utilize only the latent representation $z = E(S_t^{\text{motion}})$ for downstream tasks. To mitigate distribution shifts caused by early random exploration , we fine-tune the encoder using a replay buffer containing equal samples of mocap and simulated trajectories. Once the simulated style distribution converges to the reference, encoder fine-tuning is terminated to conserve resources.

\subsubsection{Motion Predictor}\label{ssub:mpred}
The motion predictor estimates the next target frame $P_{ref}(Z)$ by predicting the future motion sequences based on the current motion style by taking a latent space vector $z_t$ as input and outputs $\hat{S}_{t+1}^{\text{motion}}$.

\begin{figure}[hbt]
\centering
\includegraphics[width=0.99\linewidth,trim=0 0 0 350,clip]{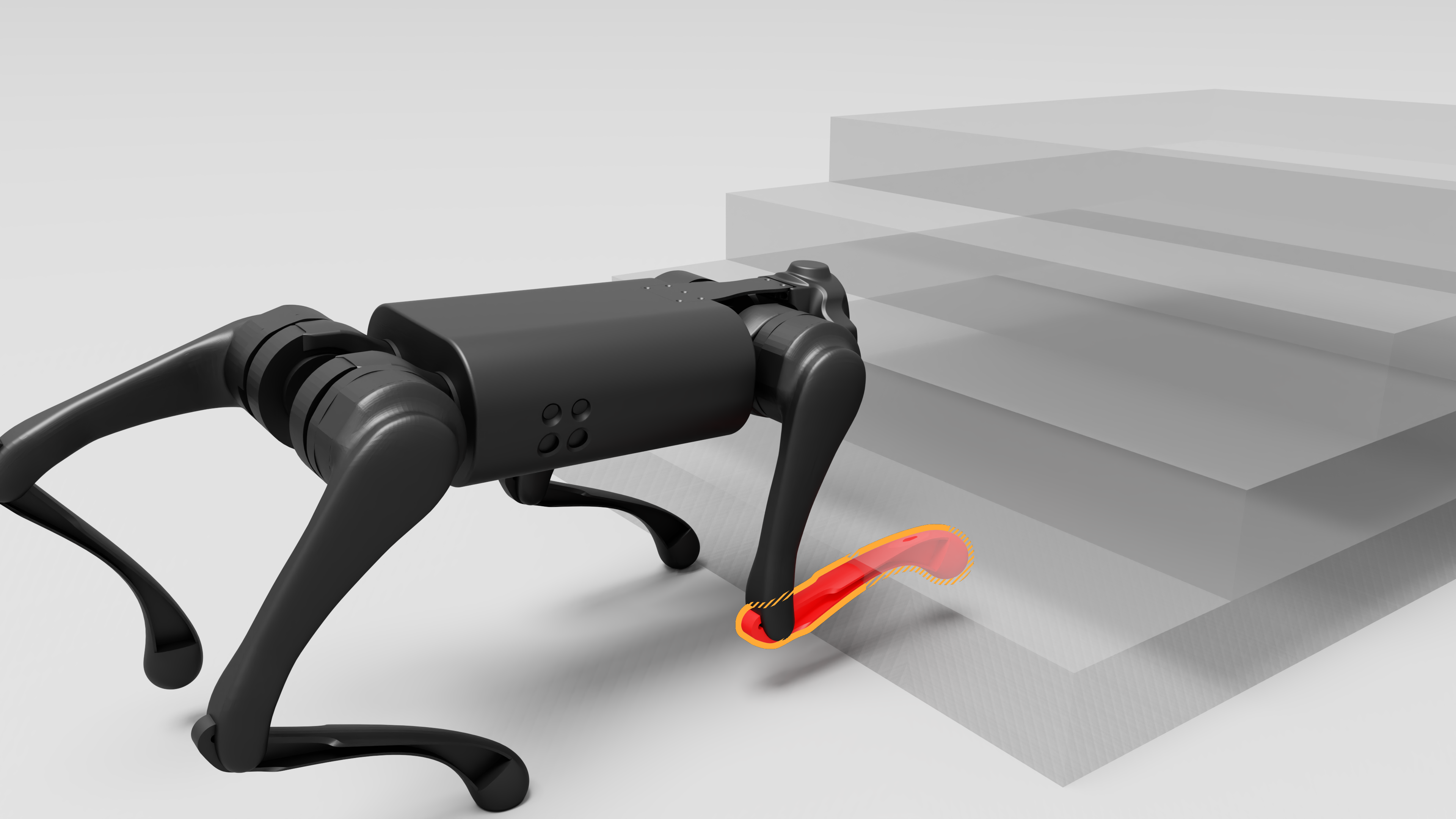}
\caption{A target motion predicted from flat terrain data may result in collisions when applied to stairs.}\label{fig:leg_intersect}\vspace{-6mm}
\end{figure}

Although this predictor performs adequately on flat terrain, it may generate invalid targets when applied to complex terrains (e.g., resulting in a leg trajectory that clips a stair; see Fig.~\ref{fig:leg_intersect}). To prevent such geometric constraint violations during policy learning, the predictor is further refined using motion frames sampled from various terrains (see Sect.~\ref{ssub:treplaybuffer}).

\subsection{Latent Space Imitation Learning}\label{sub:imitationlearning}
Given the established latent space $\mathcal{Z}$, the control policy is trained to minimize the marginal latent divergence $\mathcal{J}_{LM}$ defined in Eq. \ref{eq:kl_div}.
We formulate this task as a reinforcement learning problem: at each time step~$t$, the agent observes a state $s_t$, selects an action $a_t \sim \pi(a_t|s_t)$, and receives a reward $r_t$ alongside the next state $s_{t+1}$. A trajectory $\tau$ is represented as
\begin{equation}
    \tau = \{(s_0,a_0,r_0,s_1), (s_1,a_1,r_1,s_2), \ldots\},
\end{equation}
and our objective is to maximize the expected return
\begin{equation}
     J(\pi_\theta) = \mathbb{E}_{\tau \sim \pi_\theta}\left[\sum_{t} r_t\right],
\end{equation}
where $\theta$ denotes the parameters of the policy.

\begin{figure}[hbt]
\centering
\includegraphics[width=0.99\linewidth]{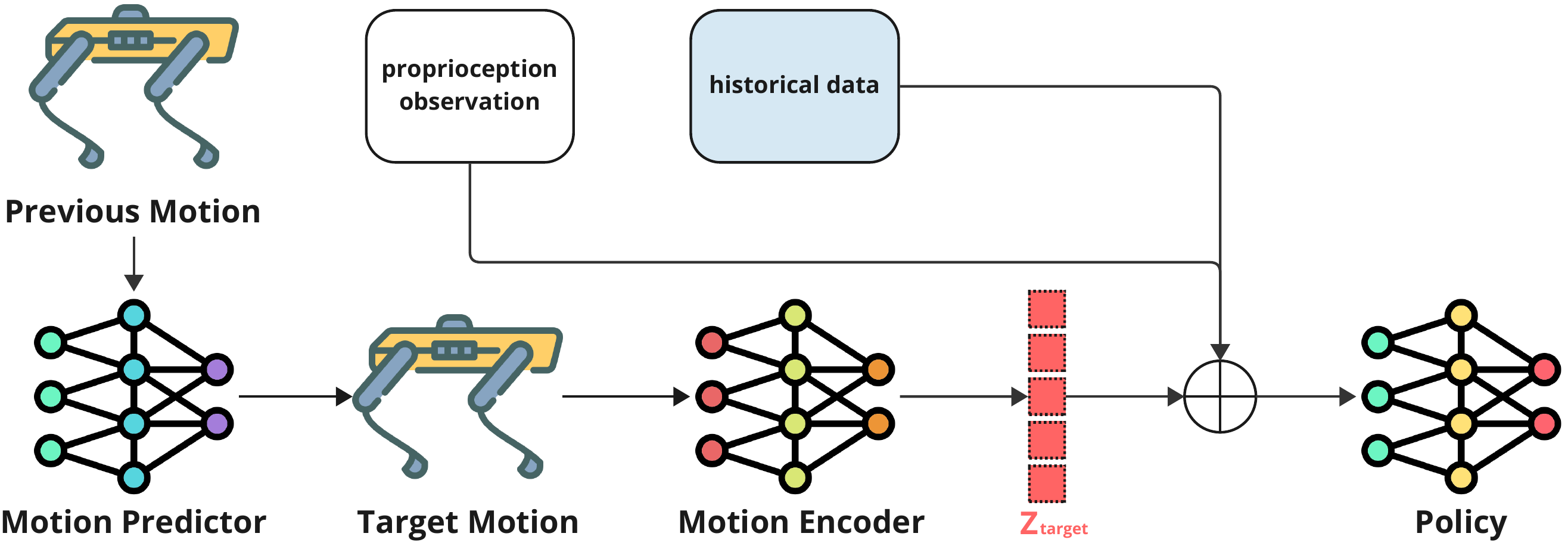}
    \caption{The input of our policy consists of proprioceptive observation, historical data, and encoded target motion $z_{\text{target}}$.}
    \label{fig:policy}\vspace{-4mm}
\end{figure}

The policy network is implemented as a three-layer feedforward neural network that accepts the state (see Fig.~\ref{fig:policy})
\[
s_t = \left(o_t^{\text{prop}},\, o_t^{\text{history}},\, o_t^{\text{target}}\right),
\]
where:
\begin{enumerate}
    \item $o_t^{\text{prop}} = (v_t,\dot{q}_t, g_t)$ represents proprioceptive observations (linear/angular velocity, projected gravity);
    \item $o_t^{\text{history}} = \left(q^{\text{imu}}_{t-H:t},\, a_{t-H:t},\, q_{t-H:t},\, \dot{q}_{t-H:t}\right) $ denotes the historical data over $H$ frames (IMU readings, previous actions, joint angles, and joint velocities);
    \item $o_t^{\text{target}} = z_{\text{target}}$ provides the target latent motion style, encoded from the next $N$ frames predicted by the motion predictor. Motion commands such as speed and direction are encoded inside the latent feature $z_{\text{target}}$.
\end{enumerate}

\subsubsection{Latent Mimic Reward}\label{ssub:mimicr}
To empirically optimize the analytical objective $\mathcal{J}_{LM}(\pi)$, we propose the Latent Mimic Reward. Rather than relying on heuristic multiple-attribute offsets~\cite{Peng2020-qm} or unstable adversarial min-max games~\cite{Escontrela2022-hz}, we compute the explicit distance metric directly in the latent space: 
\begin{equation}
    r_{\text{mimic}} = \exp\left[-w_{r}\,\mathbb{D}_{KL}\left(z_{\text{target}} \parallel z_{\text{sim}}\right)\right],
\end{equation}
where $z_{\text{target}}$ and $z_{\text{sim}}$ denote the latent features of the target and simulated motions, respectively. Both latent features are Gaussian distributions parameterized by the motion encoder in Sect.~\ref{ssub:mencoder}, and $w_{r} = 0.01$ is a scaling weight.

This reward formulation corresponds to the marginal latent divergence objective $\mathcal{J}_{LM}$ established in Sect.~\ref{sub:motivation}. By evaluating the imitation error solely in the latent space,  we eliminate the intricate human-designed reward employed in other works, e.g., \cite{Rudin2021-qd, Margolis2022-kc, Peng2020-qm}.

\begin{figure}[hbt]
\centering
\includegraphics[width=0.99\linewidth]{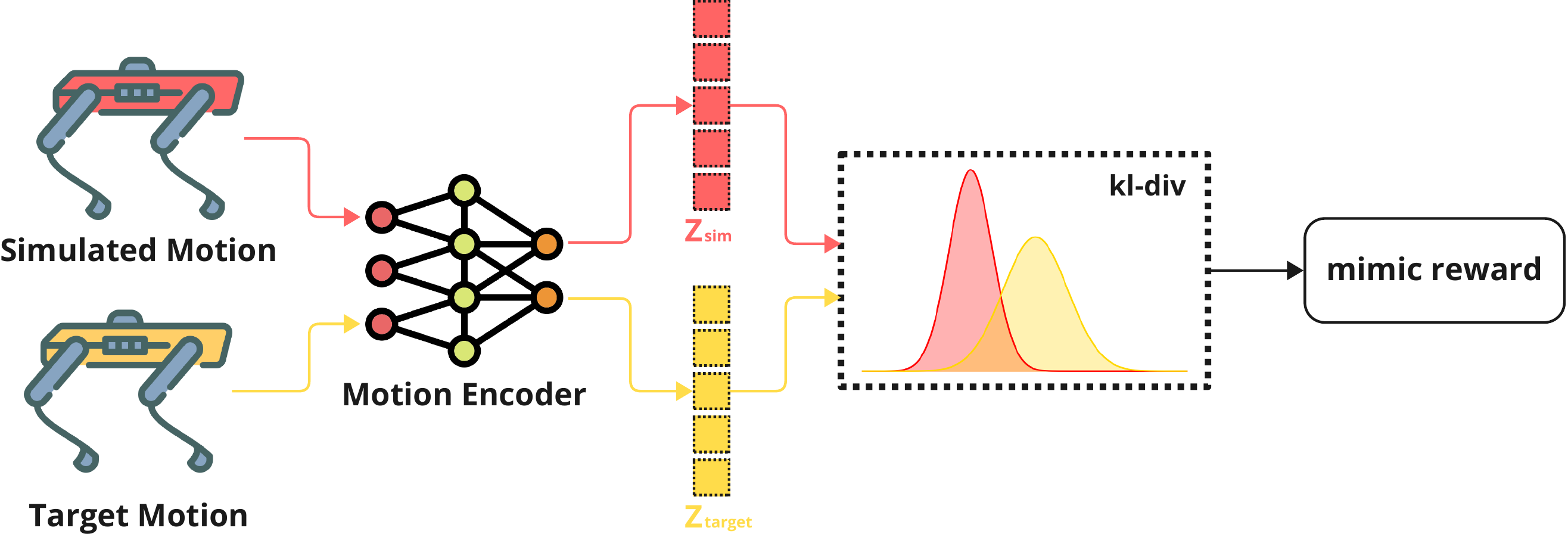}
    \caption{The latent mimic reward: both the simulated and the predicted target motions are encoded into the latent space, and their KL-divergence is computed.}
    \label{fig:latentprior}\vspace{-4mm}
\end{figure}

\subsubsection{Termination Curriculum of Mimic Tolerance}\label{ssub:mimictol}
Early termination is critical for enhancing training efficiency and mitigating the accumulation of low-quality trajectories outside the policy's support distribution~\cite{Peng2020-qm, Peng2016-aq, heess2016learning}. Large mimic errors during initial training can severely hinder policy improvement. We define the joint error for a robot with $N$ joints as
\begin{equation}
   E_{\text{joint}} = \max_{i=1,\ldots,N}\left|p_i - \hat{p}_i\right|,
\end{equation}
where $p_i$ is the simulated joint angle and $\hat{p}_i$ is the corresponding target. An episode is terminated when $E_{\text{joint}}$ exceeds a threshold $T$, which is initially set to a low value (e.g., $0.5\ \text{rad}$) and is gradually increased (up to $2\pi$) as training progresses. This termination curriculum enforces strict imitation early in training while fully engaging the conditional relaxation to provide the agent with the necessary flexibility for terrain adaptation later.

At this stage, by imitating the provided motion capture data, our method learns the locomotion controller on flat terrain as a style policy $\pi_{\theta_{\text{style}}}$ for the subsequent stage. The policy can be learned using only a single latent mimic reward (see Sect.~\ref{subsec:exp_multi_styles} for details).

\subsection{Terrain Adaptation Learning}\label{sub:terrainadapt}
As established in Sect.~\ref{sub:motivation}, truncating the conditional geometric divergence permits necessary deviations in the motion reference $\mathcal{X}$ to satisfy contact constraints on unstructured terrains. However, altering foot clearance or posture inevitably induces shifts in the state distribution relative to the flat-ground training distribution. To ensure robust traversal under these shifts without degrading stylistic fidelity, we introduce a terrain adaptation stage.

\subsubsection{Terrain Task Rewards}
In this stage, the mimic reward is augmented by incorporating a task reward~$r_{\text{task}}$ and a style-anchor-reward $r_{\text{anchor}}$.

The \textbf{Task Reward} is defined as
\begin{equation}\label{eq:task_r}
    r_{\text{task}} = r_{\text{speed}} + p_{\text{orientation}} + p_{\text{angular}},
\end{equation}
where the velocity reward is defined as $r_{\text{speed}} = w_{\text{speed}} |v - \hat{v}|$, with $v$ and $\hat{v}$ denoting the simulated and target base velocities. The orientation penalty $p_{\text{orientation}} = w_{\text{orientation}}(G_x^2 + G_y^2)$ penalizes the $x$ and $y$ components of the projected gravity vector. The angular velocity penalty is defined as $p_{\text{angular}} = w_{\text{angular}}(\omega_x^2 + \omega_y^2)$.

\textbf{Style Anchor Reward:} To prevent the policy from abandoning the learned topological gait structure in favor of greedy traversal strategies, we define the style anchor reward as
\begin{equation}\label{eq:r_anchor}
    r_{\text{anchor}} = \exp\left[w_{\text{anchor}}\,\mathbb{D}_{KL}\left(\pi_\theta(a|o) \parallel \pi_{\theta_{\text{style}}}(a|o)\right)\right],
\end{equation}
where $\pi_\theta$ represents the current policy and $\pi_{\theta_{\text{style}}}$ denotes the policy obtained during the imitation phase. The scaling weight $w_{\text{anchor}}$ balances the contribution of this reward.

The style-anchor reward is designed to preserve the motion style established during imitation learning while the robot adapts to terrain. Without this constraint, the policy tends to deviate from the original style to prioritize successful traversal. To mitigate this drift, we adopt the imitation-learning policy as an anchor policy and encourage the newly trained policy to produce an output distribution that remains consistent with it.

\subsubsection{Terrain Adaptation Replay Buffer}\label{ssub:treplaybuffer}
\begin{figure}[hbt]
\centering
\includegraphics[width=.99\linewidth]{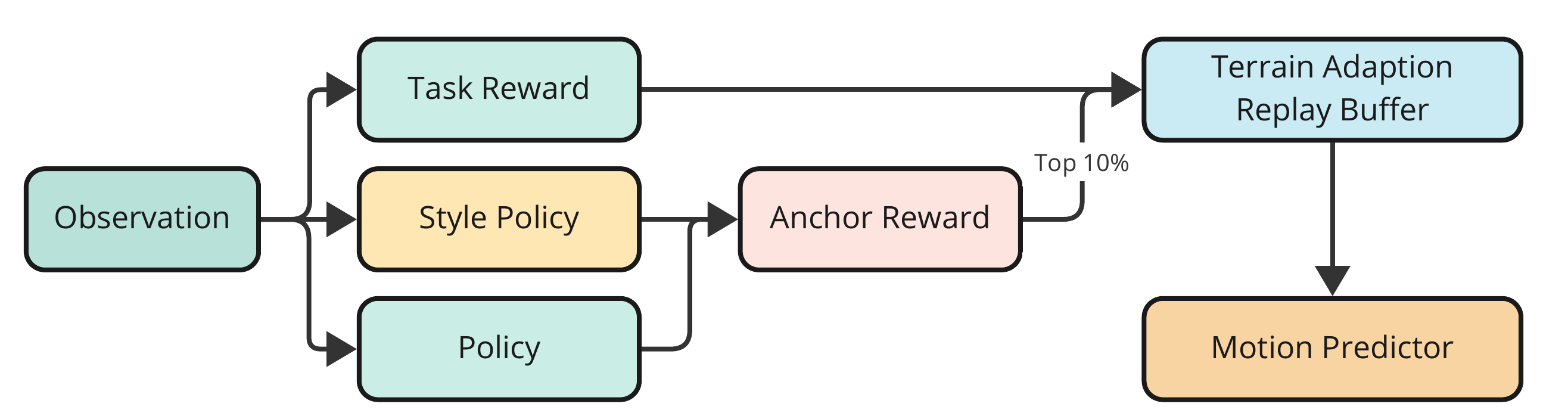}
    \caption{Terrain Adaptation Module: the current policy $\pi_{\theta}$ and the style policy $\pi_{\theta_{\text{styles}}}$ takes the same observation to output the action distribution. Transitions with top $10\%$ rewards are stored to fine-tune the motion predictor.  }
    \label{fig:terrain_adaption}\vspace{-6mm}
\end{figure}
To systematically expand the policy's adaptation domain, the robot is trained across a procedurally generated terrain curriculum. The curriculum comprises continuous difficulty levels (detailed parameters for each terrain type are quantified in Sect.~\ref{subsec:exp_terrain_adapt}).

To address state-distribution shifts induced by different terrain types, we dynamically update the support set of the latent prior. In addition to the reinforcement learning replay buffer, we maintain an additional terrain adaptation replay buffer to further enhance the motion predictor’s performance across varied terrains. As the agent progresses through the terrain curriculum, the top 10\% of transitions (as evaluated by the reward) are stored in the buffer. This curated dataset captures effective locomotion styles across diverse terrains, thereby enabling the motion predictor to better forecast the subsequent target frame under different environmental conditions (see Fig. \ref{fig:terrain_adaption}).
\section{Experiments}\label{sec:exp}

We evaluate the LatentMimic framework through comprehensive quantitative and qualitative analyses. The experiments are designed to address two primary objectives: (1) assessing stylistic fidelity during flat-ground imitation, and (2) quantifying the robustness and traversal success rates across procedurally generated irregular terrains.

\textbf{Experimental Setup and Baselines:} All policies are trained using PPO~\cite{schulman2017proximal} in Isaac Gym~\cite{makoviychuk2021isaac} with 4,096 parallel environments, a 200~Hz physics simulation, and a 50~Hz control frequency. Computations utilize a single CPU core to invoke an NVIDIA RTX 4090 GPU (24~GB), requiring $\sim$4~hours for style imitation and $\sim$12~hours for terrain adaptation. The motion encoder/predictor and actor/critic networks are parameterized as Multi-Layer Perceptrons (MLPs) with hidden dimensions of $[256, 128]$ and $[512, 256, 128]$, respectively, all employing ELU activations~\cite{clevert2016fastaccuratedeepnetwork}. The actor outputs Gaussian distributions over target joint angles for low-level PD tracking. To ensure sim-to-real transfer, domain randomization is applied uniformly at initialization (Tab.~\ref{tab:domain_rand}). We benchmark against LIA~\cite{Peng2020-qm} and AMP~\cite{alej2022adversarial}, both rigorously re-implemented on the Unitree Go1 model using their official control configurations to ensure a fair comparison of algorithmic performance.

\begin{table}[htbp]
    \centering
    \begin{tabular}{lc}
    \toprule
    \textbf{Parameter} & \textbf{Randomization Range} \\
    \midrule
    Terrain Friction        & $[0.5, 1.25]$ \\
    Additional Base Mass    & $[-1.0\ \text{kg}, 1.0\ \text{kg}]$ \\
    COM Displacement        & $[-0.15\ \text{m}, 0.15\ \text{m}]$ \\
    Motor Strength Factor   & $[0.9, 1.1]$ \\
    $K_P$ Gain Factor       & $[0.8, 1.3]$ \\
    $K_D$ Gain Factor       & $[0.5, 1.3]$ \\
    Observation Latency     & $0.03\ \text{s}$ \\
    \bottomrule
    \end{tabular}
    \caption{Domain Randomization Parameters applied during the training process to ensure sim-to-real robustness.}
    \label{tab:domain_rand}
\end{table}\vspace{-6mm}

\subsection{Multiple Locomotion Styles Imitation}\label{subsec:exp_multi_styles}

\begin{table}[htbp]
    \centering
    \setlength{\tabcolsep}{4pt} 
    \begin{tabular}{l c c c c c}
    \toprule
    \textbf{Metric} & \textbf{Method} & \textbf{Pace} & \textbf{\makecell{Pace\\Backwards}} & \textbf{Trot} & \textbf{\makecell{Trot\\Backwards}} \\
    \midrule
    \multirow{3}{*}{\makecell{Base Position\\$(\text{m}) \downarrow$}}
        & Our & $\mathbf{0.0338}$ & $\mathbf{0.0734}$ & $\mathbf{0.1437}$ & $\mathbf{0.3445}$ \\ 
        & LIA & 0.5838 & 0.3523 & 0.3519 & 0.6876 \\  
        & AMP & 0.4684 & 0.4547 & 1.4965 & 4.9182 \\  
    \midrule
    \multirow{3}{*}{\makecell{Joint Angles \\ $(\text{rad}) \downarrow$}}
        & Our & 0.1386 & $\mathbf{0.1061}$ & $\mathbf{0.1474}$ & $\mathbf{0.1511}$ \\  
        & LIA & 0.2712 & 0.1624 & 0.3611 & 0.1786 \\  
        & AMP & $\mathbf{0.0842}$ & 0.1772 & 0.1952 & 0.1702 \\  
    \midrule
    \multirow{3}{*}{\makecell{Joint Velocity\\ $(\text{rad/s}) \downarrow$}}
        & Our & $\mathbf{9.56}$ & $\mathbf{4.22}$ & $\mathbf{15.98}$ & $\mathbf{7.411}$ \\  
        & LIA & 40.45 & 15.62 & 19.40 & 19.56 \\  
        & AMP & 14.31 & 15.73 & 28.59 & 28.29 \\  
    \bottomrule
    \end{tabular}
    \caption{Tracking errors of four different locomotion styles on flat terrain. The best performance is in bold. Errors are measured using mean squared errors (MSE). Joint angle and velocity errors are computed per joint.}
    \label{tab:mimic_plane} \vspace{-2mm}
\end{table}
We perform imitation learning on four locomotion styles: pace, pace backwards, trot, and trot backwards. We use the same motion capture data as LIA and AMP, originating from~\cite{Zhang2018-dd}. Pacing refers to a motion that walks relatively slower (around $0.9\ \text{m/s}$) while moving its legs laterally (e.g, left front and left hind together). 

\begin{figure}[hbt]
\centering
\includegraphics[width=0.99\linewidth,trim=0 50 0 250,clip]{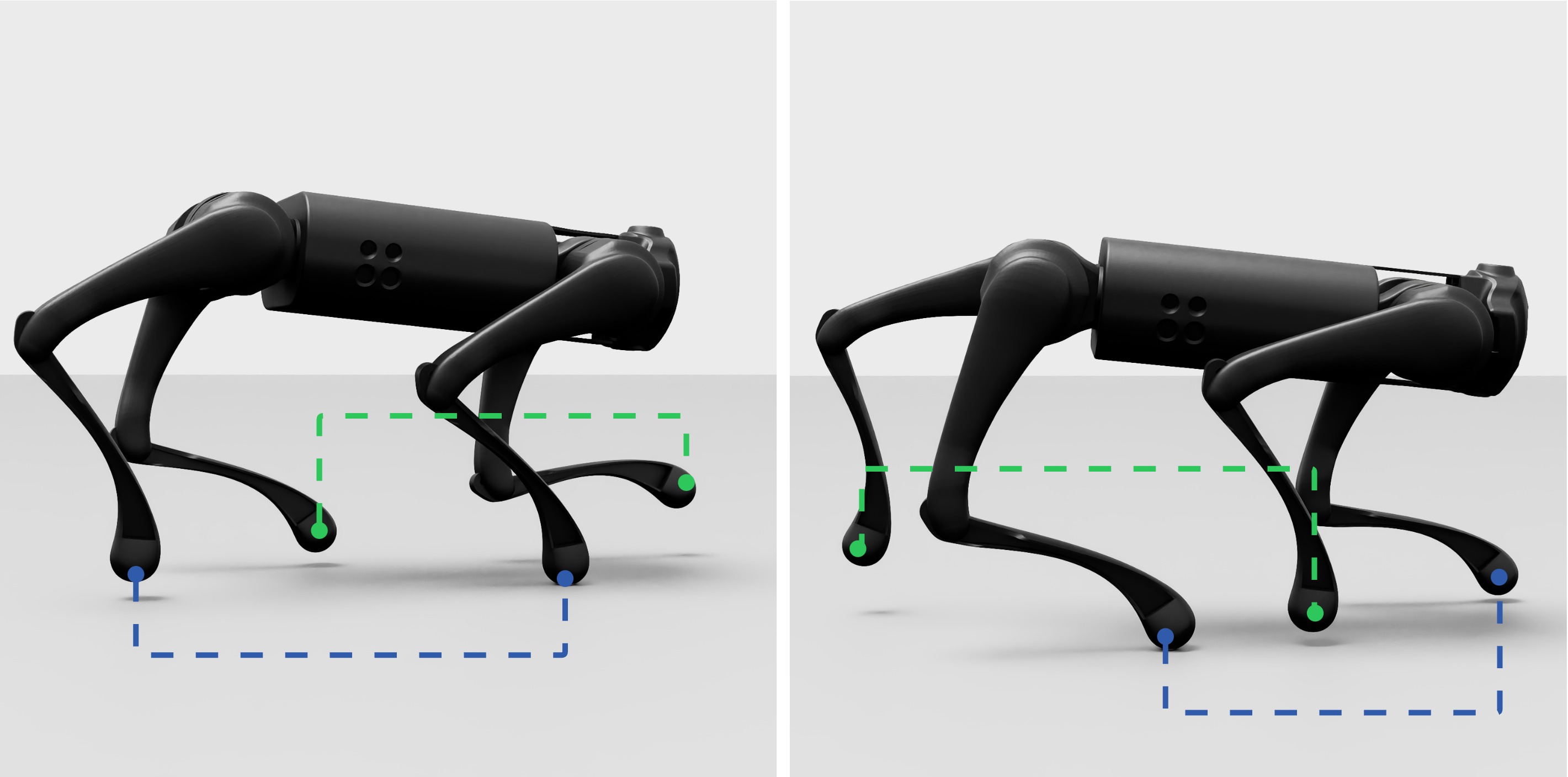}
\caption{The left figure depicts a frame of pace in which the robot moves its legs laterally (left front and left hind together). The right figure shows a trotting frame in which the robot moves its legs diagonally (left front and right hind).}\label{fig:mocap}\vspace{-6mm}
\end{figure}

Trotting moves its legs forward in diagonal pairs (e.g., left front and right hind) with a faster gait (around $1.6m/s$). We show the typical dynamics of pace and trot in Fig.~\ref{fig:mocap}. The pace backwards and trot backwards motions are generated by reversing the motion of pace and trot.

Tab.~\ref{tab:mimic_plane} reports the similarity of four different locomotion styles on terrain by calculating the mean square errors of root position, joint angles, and joint velocities of all steps. The results demonstrate that our latent mimic method outperforms all locomotion styles in minimizing the base position error. This indicates that it effectively learns to complete the walking forward task rather than merely imitating the target motion frames. Compared to both LIA and AMP, our method achieves lower errors in all three metrics except for the pace joint angles error, demonstrating the effectiveness of our Latent Space Imitation Learning approach.  

\begin{figure}[hbt]
\centering
\includegraphics[width=0.99\linewidth]{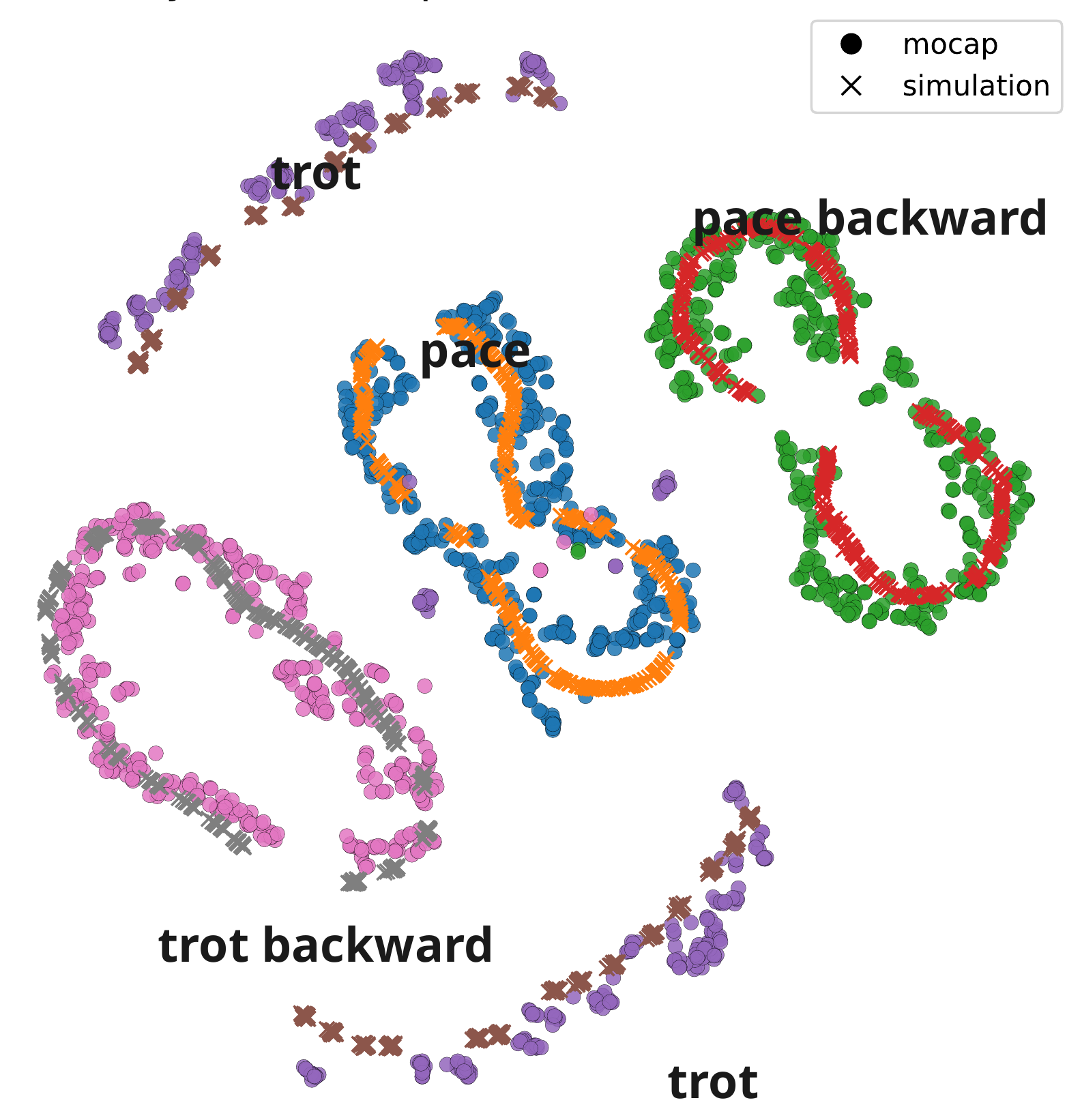}
\caption{The figure shows the t-SNE visualization of latent motion features of four different motion styles. 
}\label{fig:motfeat}\vspace{-4mm}
\end{figure}

Given that the perceived similarity between target and simulated motion styles is inherently subjective, we analyze the learned representations by projecting motion features into the latent space and visualizing them with t-SNE (see Fig.~\ref{fig:motfeat}). The visualization shows that the simulated motions closely align with the motion features derived from the mocap data across all motion categories, demonstrating that the learned policy effectively reproduces the stylistic characteristics of the reference motions across all motion styles.

\subsection{Multiple Terrain Adaptation}\label{subsec:exp_terrain_adapt}

\begin{table}[h!]
    \centering
    \begin{subtable}{\linewidth}
        \centering
        \begin{tabular*}{\linewidth}{@{\extracolsep{\fill}} w{c}{1.6cm}|w{c}{.6cm}|w{c}{.8cm}|c|w{c}{0.8cm}|c}
        \hline
         Success Rate && Pace & \makecell{Pace \\  Backwards} & Trot &  \makecell{Trot \\ Backwards} \\
        \hline
        \multirow{3}{*}{ $\geq 95\%$}
            & Our & $\mathbf{6}$ & 0 & 4 & 0 \\ 
            & LIA & 0 & 0 & 4 & 2 \\  
            & AMP & 3 & $\textbf{5}$ & $\textbf{5}$ & $\textbf{7}$ \\  
        \hline
        \multirow{3}{*}{$\geq 90\%$}
            & Our &  $\mathbf{9}$ & 1 & 5 & 3 \\  
            & LIA & 0 & 0 & $\mathbf{11}$ & 3 \\  
            & AMP & 3 & $\mathbf{5}$ & 5 & $\mathbf{7}$ \\  
        \hline
        \multirow{3}{*}{$\geq 75\%$}
            & Our & $\mathbf{22}$ & $\mathbf{18}$ & 12 & $\mathbf{8}$ \\  
            & LIA & 1 & 6 & $\mathbf{17}$ & 5 \\  
            & AMP & 3  & 5 & 5 & 7 \\  
        \hline
        \multirow{3}{*}{$\geq 50\%$}
            & Our & $\mathbf{38}$ &$\mathbf{20}$ & $\mathbf{31}$ & $\mathbf{11}$\\  
            & LIA & 3 & 12 & 20 & 6 \\  
            & AMP & 5 & 5 & 7 & 7 \\  
        \hline
        \multirow{3}{*}{$\geq 10\%$}
            & Our & $\mathbf{45}$ & $\mathbf{23}$ & $\mathbf{42}$ & $\mathbf{24}$ \\  
            & LIA & 4 & 18 & 23 & 14 \\  
            & AMP & 7 & 8 & 7 & 8 \\  
        \hline
        \end{tabular*}
        \caption{Stairs}
        \label{tab:mimic_stairs}
    \end{subtable}

    \vspace{1mm} 
  \begin{subtable}{\linewidth}
        \centering
        \begin{tabular*}{\linewidth}{@{\extracolsep{\fill}} w{c}{1.6cm}|w{c}{.6cm}|w{c}{.8cm}|c|w{c}{0.8cm}|c}
        \hline
         Success Rate && Pace & \makecell{Pace \\  Backwards} & Trot &  \makecell{Trot \\ Backwards} \\
        \hline
        \multirow{3}{*}{ $\geq 95\%$}
            & Our & $\mathbf{19}$ & 4 & 4 & 17 \\ 
            & LIA & 0 & 2 & $\mathbf{6}$ & 12 \\  
            & AMP & 9 & $\mathbf{17}$ & 9 & $\mathbf{19}$ \\  
        \hline
        \multirow{3}{*}{$\geq 90\%$}
            & Our & $\mathbf{19}$ & 10 & 5 & 21 \\  
            & LIA & 0 & 3 & 8 & $\mathbf{23}$ \\  
            & AMP & 9 & $\mathbf{19}$ & $\mathbf{11}$ & 20 \\  
        \hline
        \multirow{3}{*}{$\geq 75\%$}
            & Our & $\mathbf{30}$ & 17 & 12 & 33 \\  
            & LIA & 12 & 6 & $\mathbf{16}$ & $\mathbf{34}$ \\  
            & AMP & 11  & $\mathbf{19}$ & 13 & 23 \\  
        \hline
        \multirow{3}{*}{$\geq 50\%$}
            & Our & $\mathbf{55}$ & $\mathbf{23}$ & $\mathbf{31}$ & 40\\  
            & LIA & 25 & 8 & 18 & $\mathbf{41}$ \\  
            & AMP & 13 & 23 & 15 & 27 \\  
        \hline
        \multirow{3}{*}{$\geq 10\%$}
            & Our & $\mathbf{64}$ & 26 & $\mathbf{41}$ & $\mathbf{64}$ \\  
            & LIA & 31 & 19 & 25 & $\mathbf{64}$ \\  
            & AMP & 15 & $\mathbf{33}$ & 19 & 29 \\  
        \hline
        \end{tabular*}
        \caption{Waves}
        \label{tab:mimic_waves}
    \end{subtable}

    \vspace{1mm}

   \begin{subtable}{\linewidth}
        \centering
        \begin{tabular*}{\linewidth}{@{\extracolsep{\fill}} w{c}{1.6cm}|w{c}{.6cm}|w{c}{.8cm}|c|w{c}{0.8cm}|c}
        \hline
         Success Rate && Pace & \makecell{Pace \\  Backwards} & Trot &  \makecell{Trot \\ Backwards} \\
        \hline
        \multirow{3}{*}{ $\geq 95\%$}
            & Our & $\mathbf{10}$ & $\mathbf{20}$ & $\mathbf{28}$ & 20 \\ 
            & LIA & 0  & 3  & 15 & 4 \\  
            & AMP & 2 & 1 & 14 & $\mathbf{27}$ \\  
        \hline
        \multirow{3}{*}{$\geq 90\%$}
            & Our & $\mathbf{19}$ & $\mathbf{22}$ & $\mathbf{40}$ & $\mathbf{39}$ \\  
            & LIA & 1  & 6  & 27 & 8 \\  
            & AMP & 2 & 1 & 17 & 28 \\  
        \hline
        \multirow{3}{*}{$\geq 75\%$}
            & Our & $\mathbf{35}$ & $\mathbf{48}$ & $\mathbf{46}$ & $\mathbf{43}$ \\  
            & LIA & 32 & 10 & 32 & 33 \\  
            & AMP & 4 & 1 & 19 & 33 \\  
        \hline
        \multirow{3}{*}{$\geq 50\%$}
            & Our & 46 & $\mathbf{54}$ & $\mathbf{64}$ & 49\\  
            & LIA & $\mathbf{58}$ & 32 & 58 & $\mathbf{52}$ \\  
            & AMP & 7 & 1 & 22 & 37 \\  
        \hline
        \multirow{3}{*}{$\geq 10\%$}
            & Our & $\mathbf{64}$ & $\mathbf{64}$ & $\mathbf{64}$ & $\mathbf{64}$ \\  
            & LIA & $\mathbf{64}$ & $\mathbf{64}$ & $\mathbf{64}$ & $\mathbf{64}$ \\  
            & AMP & 13 & 8 & 28 & 37 \\  
        \hline
        \end{tabular*}
        \caption{Noise}
        \label{tab:mimic_noise}
    \end{subtable}

    \caption{Robots' success rate of traversing different terrains. Each row shows the maximum level the robot can reach with a specific success rate. Note: All methods achieve a 100\% success rate on flat terrain.}
    \label{tab:combined} \vspace{-7mm}
\end{table}

During the terrain adaptation learning phase, we evaluate our policy on three different terrains: \textit{Stairs}, \textit{Waves},  \textit{Noise}. We implemented a terrain curriculum learning method from~\cite{Rudin2021-qd} and utilized the built-in terrain curriculum in AMP. During curriculum terrain learning, we adopt a progressive training strategy: the robot starts on the easiest level and difficulty increases only after successful traversal at the current level. We compare robots' terrain adaptation by measuring the success rate at different levels in Tabs.~\ref{tab:mimic_stairs}, \ref{tab:mimic_waves}, and~\ref{tab:mimic_noise}. Since different locomotion styles exhibit varying adaptability to distinct environments, we evaluate their performance across three terrain types. The difficulty of the structured terrains scales linearly with a discrete level ranging from 1 to 64.
\squishlist
    \item \textbf{Stairs:} The stair width (run) is fixed at 0.3 m. The stair height (rise) increases proportionally with the difficulty level, scaling from 0.05 m at level 1 to a maximum of 0.23 m at level 64.
    \item \textbf{Waves:} This terrain consists of five consecutive waves. The wave amplitude scales with the difficulty level, increasing from 0.0 m at level 1 to 0.2 m at level 64. This corresponds to a maximum peak-to-valley height difference of 0.4 m at the highest level.
    \item \textbf{Noise:} The terrain is generated by sampling heights from a uniform distribution on an 80 $\times$ 80 resolution grid. The sampling bounds scale linearly with the difficulty level, expanding from $[0.0, 0.0]$ m at level 1 to $[-0.1, 0.1]$ m at level 64.
\squishend
 
We observed that our method achieves the highest terrain level in all locomotion styles across the three terrains, except for the pace backwards on waves, demonstrating its effectiveness in terrain adaptation and locomotion style preservation. Additionally, the AMP method exhibits favorable performance at the initial levels, but the success rate significantly declines after a certain number of levels. 
This is because the AMP mimic reward guides the imitation of motion styles effectively. However, due to limited terrain-adapted motion reference sampled from different terrains, AMP's ability to adapt to different terrains diminishes after the initial levels compared to our approach.

\subsection{Terrain Adaptation Module Ablation}
The terrain adaptation module is key to enhancing robots' locomotion ability over diverse terrains. We conduct an ablation study to assess the effectiveness of our model by comparing the maximum level it can achieve with and without the terrain adaptation module (Tab.~\ref{tab:abla_ta}). The result indicates that with the terrain adaptation module, the model achieves $+29$ and $+21$ additional levels on the stair and wave terrains, respectively. Additionally, both configurations can successfully reach all levels on the noise terrain, as the noise terrain is a relatively easy setup.
\begin{table}[hbt]
    \centering
    \begin{tabular}{l|c|c}
    \hline
    Pace max level & our & w/o TA \\
    \hline
    Stairs & 45 & 18 \\
    Noise & 64 & 64 \\
    Waves & 64 & 43 \\
    \hline
    \end{tabular}
    \caption{Ablation Study: Our model can achieve the maximum level with and without the terrain adaptation module for locomotion style pace.}
    \label{tab:abla_ta}\vspace{-6mm}
\end{table}



\section{Conclusions}
We introduced a novel learning framework enabling terrain-adaptive locomotion policy while adhering to target locomotion styles. This framework employs latent space imitation learning and terrain adaptation to achieve this goal. The learning process comprises two stages: the first trains a style policy using a single reward, and the second transfers the learned locomotion styles across various terrains.

However, the effectiveness of our approach depends on the quality of the motion encoder and motion predictor, necessitating an additional training process. Furthermore, this method weakens the command-control input, leading to a closer correlation between the motion velocity and direction and the motion capture data than in manually designed reward methods and AMP. Additionally, the availability of motion capture data limits the terrain adaptation capability. As illustrated in the experiments, certain locomotion styles are unsuitable for specific terrains due to inherent factors. In the future, we intend to extend our work by incorporating diverse input sources, such as video clips or human-crafted target frames, to enable richer learning of locomotion styles.

\bibliographystyle{ieeetr}
\bibliography{reference}        

\end{document}